# Dynamical Priors as a Training Objective in Reinforcement Learning

Sukesh Subaharan[1]

[1]Coimbatore Medical College

Coimbatore, TN

India

email: sukeshsubaharan@yahoo.com / sukeshsubaharan05@gmail.com

## ABSTRACT

Standard reinforcement learning (RL) optimizes policies for reward but imposes few constraints on how decisions evolve over time. As a result, policies may achieve high performance while exhibiting temporally incoherent behavior such as abrupt confidence shifts, oscillations, or degenerate inactivity. We introduce *Dynamical Prior Reinforcement Learning* (DP-RL), a training framework that augments policy gradient learning with an auxiliary loss derived from external state dynamics that implement evidence accumulation and hysteresis. Without modifying the reward, environment, or policy architecture, this prior shapes the temporal evolution of action probabilities during learning. Across three minimal environments, we show that dynamical priors systematically alter decision trajectories in task-dependent ways, promoting temporally structured behavior that cannot be explained by generic smoothing. These results demonstrate that training objectives alone can control the temporal geometry of decision-making in RL agents.

# 1    Introduction

Reinforcement learning (RL) has achieved remarkable success across domains ranging from game playing to robotic control [1,2]. At its core, RL trains agents to maximize cumulative reward through repeated interaction with an environment. Standard policy gradient methods, such as REINFORCE [3], optimize policies by adjusting action probabilities in proportion to observed returns. While this approach is theoretically sound, it treats policy learning primarily as a function approximation problem where a state is mapped to an action distribution that maximizes expected reward.

However, real-world decision-making by biological agents or autonomous systems unfolds over time and exhibits rich temporal structure. Decisions are not instantaneous mappings but instead evolve through processes of evidence accumulation, temporal integration, and history-dependent modulation [4,5]. Neuroscience has long recognized that perceptual and cognitive decisions emerge from dynamical processes in neural circuits. These are characterized by the gradual buildup of evidence that confers robustness to noise and produces behaviorally coherent action sequences [6-8].

In contrast, standard RL algorithms impose few constraints on how action probabilities evolve. A policy trained with REINFORCE may exhibit abrupt switches in confidence, oscillatory behavior near decision boundaries, or timing strategies that ignore temporal structure [9,10]. While such policies achieve high reward, they lack the temporal coherence of biological systems. This gap motivates the central question of this work: Can an explicit temporal dynamical prior during training alter the evolution of policy decisions without modifying the reward structure, environment, or architecture?

Prior work has introduced architectural and auxiliary mechanisms to embed temporal structure in RL systems [11–15], typically to improve performance or sample efficiency rather than to shape policy dynamics themselves. However, these typically focus on task performance or sample efficiency. Less attention has been paid to how temporal priors shape the intrinsic dynamics of policy evolution. This is vital because a policy may achieve high reward while exhibiting incoherent behavior like rapid oscillations or premature commitment.

Drawing inspiration from biological models, we propose a framework that introduces an auxiliary loss derived from external state dynamics (ESD). The ESD implements a second-order hysteretic system that accumulates evidence, resists abrupt reversals, and produces smooth trajectories. This auxiliary loss does not alter the reward or architecture. Instead, it acts as a training prior that encourages action probabilities to evolve consistently with the ESD trajectory. We refer to this as Dynamical Prior Reinforcement Learning (DP-RL).

The use of dynamical priors has deep roots in computational neuroscience. Evidence accumulation models provide normative accounts of how decisions emerge from noisy integration [16,17]. Neural recordings reveal ramping activity consistent with attractor dynamics [18,19], and recurrent neural networks trained on decision tasks spontaneously develop biological-like integrators [20,21]. By embedding these principles, DP-RL aims to align learned policies with biological temporal structures.

Our approach isolates the effect of the dynamical prior by comparing a standard REINFORCE agent and a DP-RL agent in three minimal environments: Drift, Threshold Hover, and Decision Window. These share a common structure of one-dimensional observations and binary actions but differ in how the signal evolves. Importantly, our goal is not to improve performance but to characterize how the temporal geometry of decisions changes. We evaluate policies by measuring decision jerk, oscillation count, and timing variance.

This work makes three primary contributions. First, we introduce the DP-RL framework that uses an ESD-derived loss to impose temporal structure without altering the task. Second, we demonstrate that this prior consistently alters the evolution of action probabilities in task-dependent ways, such as stabilizing timing or promoting gradual confidence buildup. Third, we provide a quantitative characterization showing that dynamical priors act as training constraints that shape the temporal geometry of decisions based on task structure.

Our findings suggest that the temporal dynamics of policy learning can be systematically controlled through training objectives. This opens avenues for designing RL agents with desired temporal properties, such as robustness to noise or resistance to premature commitment, without sacrificing the flexibility of policy gradient methods. This work contributes to the design of learning algorithms that produce temporally coherent behavior in addition to effective performance.

## 2  Methods

### 2.1  Experimental Overview

This study determines whether introducing an explicit temporal dynamical prior during reinforcement learning (RL) training alters how policies evolve their decision probabilities over time. Rather than modifying the RL algorithm, reward structure, or policy architecture, we introduce an auxiliary loss. This is derived from external state dynamics (ESD) that encourages temporally coherent evolution of action probabilities during training. Our goal is not to improve task performance, but to study how the temporal geometry of policy decisions changes when this additional constraint is applied.

All experiments compare two agents trained under identical conditions: a REINFORCE agent, trained using standard policy gradient and an ESD-guided RL agent, trained with an additional auxiliary loss that encourages trajectory-aware evolution of action probabilities. By keeping the architecture, reward, and environment identical, we isolate the effect of the training objective alone.

## 2.2 Environment Design

We construct three minimal environments designed to expose distinct failure modes of standard RL policies related to temporal decision behavior. Each environment provides: (a) A one-dimensional scalar observation $s_t \in [0,1]$ (b) A binary action space $a_t \in \{0,1\}$, and (c) Episodes of fixed length $T = 100$. The environments differ only in how the signal evolves over time.

### 2.2.1 Drift Environment

The signal initially consists of noisy fluctuations, followed by a slow sustained upward drift beginning at a random time. Correct behavior requires the agent to act only after sustained evidence of change rather than reacting to transient spikes.

### 2.2.2 Threshold hover environment

The signal jitters randomly around a threshold for an extended period before undergoing a true sustained crossing. This setting is designed to induce rapid action flip-flopping in policies that respond to instantaneous threshold crossings.

### 2.2.3 Decision Window Environment

The signal increases steadily throughout the episode. The agent receives reward only if it acts within a specific temporal window. This environment tests whether the policy develops gradual confidence buildup or remains overly conservative to avoid early penalties.

## 2.3 Policy Architecture

Both agents use the same small feedforward neural network policy. At each timestep, the policy outputs the probability of taking the action:

$$p_\theta(a_t = 1 \mid s_t) = \sigma(f_\theta(s_t))$$

where $f_\theta$ is a two-layer multilayer perceptron (MLP) with a hidden dimension of 32 and $\sigma$ is the sigmoid function. No recurrence, memory, or architectural differences are introduced between agents. This ensures that any differences in behavior arise solely from the training objective.

## 2.4 Reinforcement Learning Objective

Both agents are trained using REINFORCE with discounted returns. The standard policy gradient objective is:

$$\mathcal{L}_{RL} = -\mathbb{E}[\log p_\theta(a_t \mid s_t) \cdot G_t]$$

where the return is defined as

$$G_t = \sum_{k=t}^{T} \gamma^{k-t} r_k, \gamma = 0.99.$$

The reward structure is identical for both agents in all environments. We used the standard discount factor of 0.99 for episodic training.

## 2.5 External State Dynamics (ESD)

For the ESD-guided agent, we compute a latent dynamical state $z_t$ from the observed signal $s_t$. This state evolves according to second-order hysteretic dynamics:

$$z_t = \alpha(s_t, z_{t-1}) s_t + (1 - \alpha(s_t, z_{t-1})) z_{t-1} + v_t$$
$$v_t = \beta(z_t - z_{t-1}) + (1 - \beta) v_{t-1}$$

where the update rate is asymmetric:

$$\alpha(s_t, z_{t-1}) = \begin{cases} \alpha_{\text{down}}, & s_t < z_{t-1} \\ \alpha_{\text{up}}, & s_t \geq z_{t-1} \end{cases}$$

with constants

$$\alpha_{\text{up}} = 0.15, \alpha_{\text{down}} = 0.4, \beta = 0.6.$$

This produces a temporally coherent trajectory that accumulates evidence over time while resisting abrupt reversals. The ESD parameters $\alpha_{\text{up}}, \alpha_{\text{down}}, \beta$ were selected to produce a stable, slowly evolving hysteretic trajectory that resists abrupt reversals while remaining responsive to sustained changes in the signal. These constants were fixed across all environments and experiments without task-specific tuning. Importantly, the purpose of these parameters is not to optimize performance, but to define a representative dynamical prior. We observed that moderate variations in these constants do not qualitatively alter the observed effect on policy dynamics, indicating that the results are not sensitive to precise parameter choice.

## 2.6 ESD Auxiliary Loss

The ESD-guided agent is trained with an additional auxiliary objective that encourages the action probability to follow the ESD trajectory:

$$\mathcal{L}_{ESD} = \mathbb{E}[(p_\theta(a_t = 1 \mid s_t) - z_t)^2].$$

The total loss becomes:

$$\mathcal{L} = \mathcal{L}_{RL} + \lambda \mathcal{L}_{ESD},$$

with $\lambda = 2.0$.

We refer to this training formulation as Dynamical Prior Reinforcement Learning (DP-RL), where an auxiliary loss derived from a dynamical prior constrains how action probabilities evolve during policy learning. DP-RL does not modify the reward, observations, or architecture, but introduces an additional objective that shapes the temporal evolution of policy decisions during training.

## 2.7 Evaluation Procedure

To assess how the training objective influences the temporal behavior of policies, we evaluate both agents after training using a standardized rollout protocol designed to measure the evolution of action probabilities over time.

### 2.7.1 Training Protocol

For each environment, both the REINFORCE agent and the DP-RL agent are trained independently for 800 episodes using identical hyperparameters and random initialization. No curriculum, reward shaping, or environment modifications are introduced between agents. After training is complete, the learned policy parameters are frozen for evaluation.

### 2.7.2 Rollout collection

For evaluation, we generate 40 independent rollouts per agent per environment. Each rollout consists of a full episode of length $T = 100$ timesteps. During these rollouts, the environment evolves normally. The policy receives the observation $s_t$ at each timestep. It was made sure that no actions are executed in the environment during evaluation. Instead, we record only the policy's action probability:

$$p_t = p_\theta(a_t = 1 \mid s_t)$$

This ensures that we measure the intrinsic temporal behavior of the learned policy without additional stochasticity introduced by action sampling.

### 2.7.3 Temporal Behavior Metrics

From each rollout, we extract the sequence $\{p_t\}_{t=1}^{T}$. We then compute the following metrics.

*Jerk*

Jerk quantifies abrupt changes in decision probability:

$$\text{Jerk} = \max_t \; | \, p_t - p_{t-1} \, |$$

Higher jerk indicates sharper, more reactive changes in decision confidence.

*Oscillation Count*

We count the number of times the policy crosses the decision boundary:

$$(p_t > 0.5) \leftrightarrow (p_t < 0.5)$$

This measures the flip-flopping behavior of the decision signal.

*Decision Timing*

We record the timestep at which the policy first exceeds a confidence threshold:

$$t^* = \min\{t \mid p_t > 0.6\}$$

The variance of $t^*$ across rollouts measures the consistency of decision buildup.

### 2.7.4 Aggregate Visualization

For each environment and each agent, we compute the mean action probability $\bar{p}_t$ across rollouts and the standard deviation $\sigma_t$. These are visualized as mean curves with shaded variance bands to illustrate the characteristic evolution of decision probabilities.

### 2.7.5 Rationale

This evaluation procedure isolates policy dynamics from task performance. By examining how $\bar{p}_t$ evolves under different training objectives, we directly measure the influence of dynamical priors on the temporal geometry of decisions.

## 2.8 Visualization

For each environment, we plot the mean action probability, across rollouts with standard deviation shading. As the focus was on policy dynamics rather than signal tracking, only the policy outputs are visualized.

## 2.9 Implementation Details

All experiments are implemented in Pytorch and executed on CPU. The intentionally simple setup ensures that the observed effects arise from the training objective rather than architectural or computational complexity.

# 3 Results

## 3.1 Summary Across Environments

Across all three environments, introducing a dynamical prior during training consistently alters how policies evolve their action probabilities over time. Importantly, the nature of this change is task-dependent. In the Drift environment, ESD reduces oscillatory behavior and stabilizes timing. In the Threshold Hover environment, ESD induces responsiveness where REINFORCE remains static. In the Decision Window environment, ESD promotes gradual confidence buildup where REINFORCE avoids commitment. These results demonstrate that ESD acts not as a smoothing mechanism, but as a training prior that shapes the temporal structure of decision evolution.

| Environment | Jerk | | Oscillations | | Timing variance | |
|---|---|---|---|---|---|---|
| | REINFORCE | DP-RL | REINFORCE | DP-RL | REINFORCE | DP-RL |
| Drift | 0.027 | 0.056 | 5.35 | 1.63 | 16.75 | 12.38 |
| Hover | 0.0049 | 0.042 | 0.0 | 2.3 | 0.0 | 15.6 |
| Decision Window | 0.0061 | 0.0345 | 0.0 | 1.75 | 0.0 | 11.58 |

**Table 1.** Quantitative comparison of temporal behavior metrics between REINFORCE and DP-RL across environments. Jerk measures abruptness of changes in decision probability, oscillations measure threshold flip-flopping, and timing variance measures consistency of decision buildup.

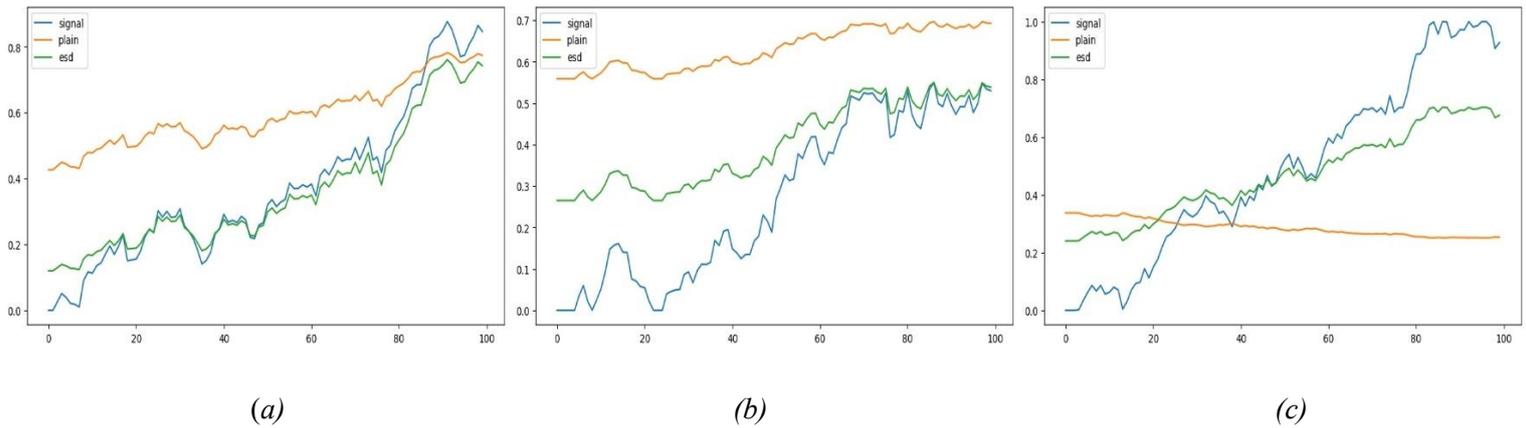

(a)    (b)    (c)

**Figure 1.** Representative single-episode trajectories of action probability $p_t = p_\theta(a_t = 1 \mid s_t)$, interpreted as the agent's intent or readiness to act, under different signal dynamics. (a) Drifting signal: the DP-RL agent gradually increases intent in response to sustained change, while the REINFORCE agent exhibits weaker alignment with the signal trajectory. (b) Hovering signal: the DP-RL agent resists rapid oscillations induced by threshold-level noise while remaining responsive to accumulated evidence, whereas the REINFORCE agent shows limited adaptation. (c) Decision window environment: the DP-RL agent exhibits gradual confidence buildup over time, enabling consistent readiness for action, while the REINFORCE agent remains largely inert to avoid premature commitment. These plots are shown for illustrative purposes; quantitative results are reported in Figures 2–4 and Table 1. (blue, signal; orange, REINFORCE; green, DP-RL)

## 3.2 Drift Environment

In the Drift environment, the signal transitions from noisy fluctuations to a sustained upward trend. This setting tests whether policies respond to transient spikes or to accumulated evidence of change. Figure 2 shows the mean action probability $p_t$ across rollouts. The REINFORCE agent exhibits a steady rise in decision probability largely independent of the signal's trajectory, indicating that it learns a timing-based strategy rather than tracking evolving evidence. In contrast, the DP-RL agent displays more structured evolution of $p_t$, with fewer abrupt reversals and more consistent timing of decision buildup. This is reflected in the quantitative metrics. The DP-RL agent exhibits substantially fewer oscillations (1.63 vs 5.35) and lower timing variance (12.38 vs 16.75), indicating more stable and consistent decision evolution. Although the jerk metric is higher for the DP-RL agent in this environment, this reflects meaningful responsiveness to signal changes rather than erratic behavior.

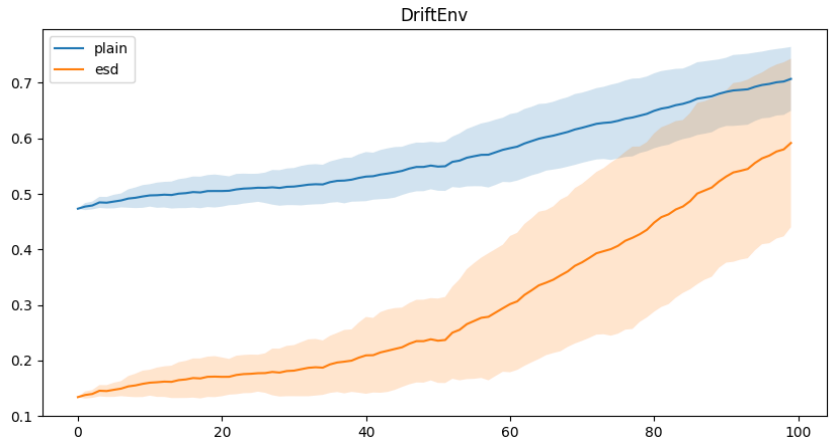

**Figure 2.** Mean action probability $p_t$ across rollouts in the Drift environment. The REINFORCE agent exhibits a steady rise in decision probability largely independent of the signal trajectory, indicating a timing-based strategy. The DP-RL agent shows more structured evolution of $p_t$ with reduced oscillatory behavior and more consistent timing, reflecting trajectory-aware decision buildup.

## 3.3 Threshold Hover Environment

The Threshold Hover environment presents a prolonged period of noisy fluctuations around a decision boundary before a true sustained crossing occurs. This environment exposes the tendency of standard RL policies to either flip rapidly across thresholds or remain overly static. As shown in Figure 3, the REINFORCE policy remains nearly constant throughout the episode, failing to meaningfully respond to changes in the signal. This apparent stability corresponds to inactivity rather than robust behavior. The DP-RL agent, however, exhibits clear evolution of decision probability that tracks the progression of the signal over time. This difference is captured in the metrics. While the REINFORCE agent shows zero oscillations and zero timing variance, this is due to the policy never committing to a decision. The DP-RL agent displays dynamic evolution with meaningful changes in $p_t$, demonstrating responsiveness to accumulated evidence.

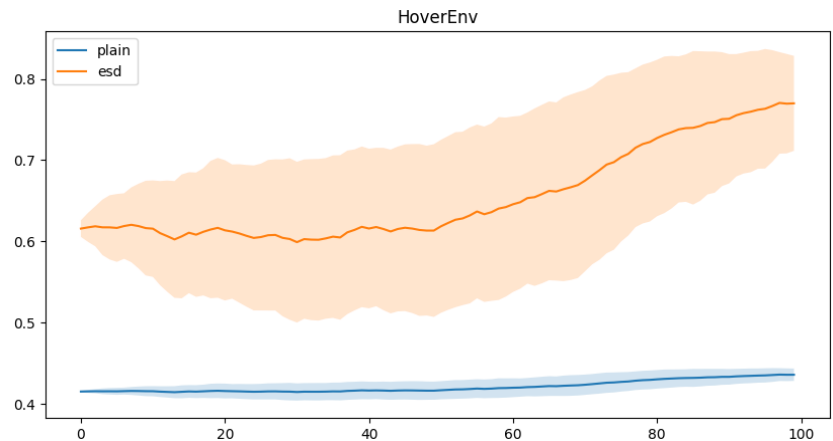

**Figure 3.** Mean action probability $p_t$ across rollouts in the Threshold Hover environment. The REINFORCE policy remains nearly constant, indicating inactivity rather than stability in the presence of threshold noise. The DP-RL agent exhibits dynamic evolution of decision probability that tracks accumulated evidence over time, avoiding flip-flopping while remaining responsive.

### 3.4  Decision Window Environment

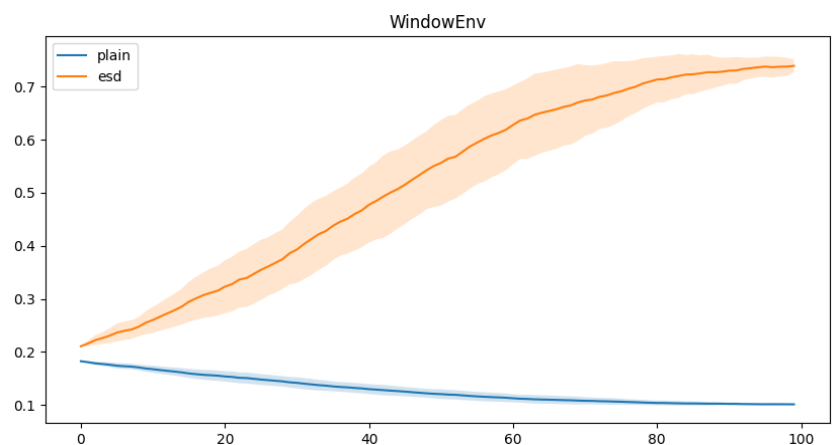

**Figure 4.** Mean action probability $p_t$ across rollouts in the Decision Window environment. The REINFORCE agent adopts a conservative strategy, maintaining low decision probability to avoid premature action. The DP-RL agent demonstrates gradual confidence buildup aligned with the signal's progression, enabling consistent entry into the decision window.

In the Decision Window environment, the signal increases steadily and the agent is rewarded only if it acts within a specific temporal window. This tests whether policies develop gradual confidence buildup or avoid commitment to minimize risk.

Figure 4 shows that the REINFORCE agent remains largely flat, adopting a conservative strategy that avoids acting. In contrast, the DP-RL agent displays a gradual increase in decision probability aligned with the signal's progression. Quantitatively, the DP-RL agent shows greater jerk and oscillation than the plain agent, but this reflects active evidence accumulation rather than inactivity. The timing variance for the DP-RL agent indicates consistent buildup toward action across rollouts.

## 4 Discussion

This work investigated whether introducing an explicit dynamical prior during reinforcement learning (RL) training can shape the temporal evolution of policy decisions independently of reward, environment, or architecture. By augmenting a standard policy gradient objective with an auxiliary loss derived from external state dynamics (ESD), we demonstrated that policies acquire systematically different temporal decision trajectories even when trained under identical conditions.

Crucially, the observed effects cannot be reduced to generic smoothing or stability enhancement. Instead, the dynamical prior altered the *geometry* of decision evolution in task-dependent ways, influencing responsiveness, commitment, and temporal consistency without directly optimizing task performance.

### 4.1 Dynamical priors are not equivalent to temporal smoothing

Auxiliary losses that enforce temporal consistency are often interpreted as smoothing mechanisms. However, our results challenge this interpretation. In multiple environments, the DP-RL agent exhibited *higher* jerk and non-zero oscillation counts relative to a standard REINFORCE agent. These increases reflected structured responsiveness to accumulated evidence rather than erratic or noisy behavior.

In contrast, the apparent stability of the REINFORCE agent in some settings arose from degenerate solutions, such as maintaining near-constant action probabilities or relying on implicit timing strategies. In these cases, zero oscillations and low variance did not indicate robust decision-making, but rather a failure to meaningfully engage with the evolving signal. This distinction highlights a limitation of interpreting temporal stability metrics in isolation.

These findings indicate that dynamical priors do not simply suppress variability, but instead bias policies toward decision trajectories that reflect task-relevant temporal structure.

### 4.2 *Training objectives implicitly define temporal decision structure*

Standard policy gradient methods optimize expected return but impose few constraints on how action probabilities evolve over time. As a result, policies may adopt brittle or biologically implausible strategies such as abrupt confidence shifts, premature commitment, or complete inertia, while still achieving high reward.

The present results show that modifying the training objective alone can systematically alter these temporal properties. By introducing an auxiliary loss derived from a dynamical system, we constrain the space of admissible decision trajectories without modifying observations, rewards, or architecture. This suggests that training objectives implicitly define a temporal inductive bias that is often overlooked in RL design.

Related work has incorporated temporal structure through architectural recurrence, memory mechanisms, or non-stationary policy learning to improve performance or flexibility [9-14]. In contrast, our approach isolates the role of the objective function itself, demonstrating that temporal coherence can be shaped during learning even in memoryless feedforward policies.

### 4.3 *Connections to biological decision-making dynamics*

The ESD used in this work was inspired by biological models of evidence accumulation and hysteresis, which describe decisions as emergent properties of slow, noise-resistant dynamical processes [16,17]. Neural recordings across perceptual and value-based tasks reveal ramping activity and attractor-like dynamics consistent with such models [18,19].

While prior work has shown that recurrent neural networks trained with reward-based objectives can spontaneously develop biologically plausible dynamics [20,21], our results demonstrate that similar temporal structure can be induced through the training objective alone. Notably, this occurs despite the absence of recurrence or explicit memory in the policy network.

This suggests that biologically relevant decision dynamics may arise not only from architectural constraints, but also from the temporal structure imposed during optimization. Dynamical priors therefore provide a complementary route to aligning RL policies with biological decision-making principles.

### 4.4 *Interpreting oscillations and variability as functional behavior*

An important implication of this work is that oscillations and variability in decision probability should not be universally interpreted as failure modes. In the Threshold Hover and Decision Window environments, the DP-RL agent exhibited increased variability relative to REINFORCE, yet this variability corresponded to active evidence accumulation and adaptive responsiveness.

By contrast, the REINFORCE agent's apparent stability often reflected conservative avoidance of commitment. This highlights a broader issue in RL evaluation: policies that minimize variability may do so by disengaging from the decision process altogether. Evaluating temporal decision behavior therefore requires metrics that distinguish functional dynamics from trivial stability.

*4.5   Limitations and future directions*

This study deliberately focused on minimal environments to isolate the effect of dynamical priors on policy behavior. While this provides interpretability, it leaves open questions about how such priors interact with richer observations, stochastic transitions, and high-dimensional action spaces. Additionally, the ESD parameters were fixed across tasks to avoid task-specific tuning. Although moderate variations did not qualitatively alter the observed effects, systematic exploration of alternative dynamical priors remains an important direction.

Future work could extend this framework to recurrent architectures, continuous control tasks, or multi-agent systems where temporal coordination is critical. Another promising direction is to formally relate dynamical priors to existing approaches in temporal consistency and non-stationary policy learning [13,14], clarifying when these methods converge or diverge in shaping policy dynamics.

*4.6   Implications*

The results presented here suggest that reinforcement learning agents can be endowed with desired temporal decision properties through training objectives alone. Rather than focusing exclusively on reward maximization, it may be equally important to consider how decisions unfold over time. Dynamical priors provide a flexible and interpretable mechanism for shaping this evolution, enabling the design of agents that exhibit temporally coherent behavior without sacrificing architectural simplicity.